\RequirePackage{fix-cm}
\documentclass[twocolumn]{svjour3}          % twocolumn
\smartqed  % flush right qed marks, e.g. at end of proof
\usepackage{graphics} % for pdf, bitmapped graphics files
\usepackage{epsfig} % for postscript graphics files
\usepackage{times} % assumes new font selection scheme installed
\usepackage{amsmath} % assumes amsmath package installed
\usepackage{amssymb}  % assumes amsmath package installed
\usepackage{bm}
\usepackage{subfigure}
\usepackage{multirow}
\usepackage{xcolor}
\usepackage{hyperref} 
\hypersetup{
	colorlinks=true,
	linkcolor=green,
	filecolor=black,      
	citecolor=blue,
	urlcolor=cyan,
}
\usepackage{booktabs} 

\graphicspath{{img/}}
\begin{document}
	
	\title{Robot in the mirror: toward an embodied computational model of mirror self-recognition
	    \thanks{KI - Künstliche Intelligenz - German Journal of Artificial Intelligence - Springer} 
	    \thanks{M.H. and V.O. were supported by the Czech Science Foundation (GA \v{C}R), project nr. 17-15697Y. P.L. was partially supported by the H2020 project Selfception (nr.  741941).}
	 }
	
	\author{Matej Hoffmann \and
	    Shengzhi Wang      \and
		Vojtech Outrata  \and
		Elisabet Alzueta    \and
		Pablo Lanillos
	}
	
	
	\institute{
	    M. Hoffmann and V. Outrata  \at
		Department of Cybernetics, Faculty of Electrical Engineering, Czech Technical University in Prague.
		\email{matej.hoffmann@fel.cvut.cz}
		\and
	    S. Wang \at
		Technical University of Munich
		\email{shengzhi.wang@tum.de}           %  \\
		\and
		E. Alzueta \at
		Center for Health Sciences, SRI International, Menlo Park, CA 94025, USA.
		\and
		P. Lanillos \at
		Donders Institute for Brain, Cognition and Behaviour, Nijmegen, the Netherlands \email{p.lanillos@donders.ru.nl}
	}
	

	\maketitle
	
	% 150 to 200 words
	\begin{abstract}
	Self-recognition or self-awareness is a capacity attributed typically only to humans and few other species. The definitions of these concepts vary and little is known about the mechanisms behind them. However, there is a Turing test-like benchmark: the mirror self-recognition, which consists in covertly putting a mark on the face of the tested subject, placing her in front of a mirror, and observing the reactions. In this work, first, we provide a mechanistic decomposition, or process model, of what components are required to pass this test. Based on these, we provide suggestions for empirical research. In particular, in our view, the way the infants or animals reach for the mark should be studied in detail. Second, we develop a model to enable the humanoid robot Nao to pass the test. The core of our technical contribution is learning the appearance representation and visual novelty detection by means of learning the generative model of the face with deep auto-encoders and exploiting the prediction error. The mark is identified as a salient region on the face and reaching action is triggered, relying on a previously learned mapping to arm joint angles. The architecture is tested on two robots with completely different face.
	\end{abstract}

	
	\section{INTRODUCTION}
	\label{sec:intro}
	The ``Turing test'' of self-awareness or self-recognition was independently developed for chimpanzees \cite{gallup1970chimpanzees} and infants \cite{amsterdam1972mirror} and consists in covertly putting a mark on the faces of the subjects, placing them in front of a mirror, and observing their reactions. \textit{Mirror self-recognition (MSR)} is often used to denote this test. The details of the mark placement, testing procedure, and assessment differ depending on the tradition \cite{bard2006self}. In infants, a spot of rouge is covertly applied alongside the infant's nose by the mother. Several behaviors may be counted as passing the test: `Touch spot of rouge', `Turns head and observes nose', `Labels self (verbal)', or `Points to self' \cite{amsterdam1972mirror}. In chimpanzees, a dye is applied to the unconscious animal and placed on \textit{two} nonvisible locations (brow ridge and opposite ear). The assessment criteria are either `Combination of changing behaviors' (decrease social responses, increase self-directed responses) or `Touches to mark' \cite{gallup1970chimpanzees}.

    The test has, in different variants, been used many times and in different species. Often it was interpreted in a binary fashion---specific species (humans, chimpanzees, orangutans, bottlenose dolphins, Asian elephants, Eurasian magpies) can pass the test and hence ``possess self-awareness'', while other species do not. In humans, it is studied in a developmental perspective: infants pass the test at around the age of 20 months. However, de Waal~\cite{deWaal2019fish} argues against such ``Big Bang'' theory of self-awareness and advocates a gradualist perspective instead. 
	
	Despite fifty years of study of MSR, little is known about the mechanisms that bring about success in the test. A notable exception is Mitchell~\cite{mitchell1993mental}, proposing two theories. In this work, we follow up on this approach, introducing more detail, and attempt to unfold the MSR phenomenon into a block diagram, listing all the necessary prerequisites and modules. Furthermore, following the synthetic methodology (``understanding by building'') \cite{PfeiferBongard2007,HoffmannPfeifer2018} and the cognitive developmental robotics approach (e.g., \cite{asada2009cognitive,cangelosi2015developmental}), we will realize MSR on a humanoid robot, adding to the efforts to understand body representations and the self by developing embodied computational models thereof using robots (see \cite{hoffmann2010body,lanillos2017enactive,kuniyoshi2019fusing,prescott2019synthetic,hafner2020prerequisites} for surveys). Understanding MSR---still a relatively low-level milestone of the development of ``self-knowledge'' in humans \cite{neisser1988five}---will specifically generate insights into the sensorimotor, ecological \cite{neisser1988five}, or minimal \cite{gallagher2000philosophical} self. 	
	
	This article is structured as follows. Section \ref{sec:rel_work} describes the mirror mark test, the possible mechanisms behind MSR, how humans recognize their own face, and summarizes previous works on self-recognition in robotics. Section \ref{sec:mechanisms} discusses in detail the mechanisms involved in MSR and presents a process model. In Section \ref{sec:model}, we describe the methods needed for implementation on a humanoid robot---learning the appearance representation and detecting the mark in particular. Section \ref{sec:results} demonstrates quantitative and qualitative performance of the architecture components and the robot behavior under MSR on two versions of the robot Nao. Discussion is followed by Conclusion and future work.
	
    \section{RELATED WORK}
    \label{sec:rel_work}
	
	\subsection{The mirror mark test}
	\label{sec:mirror_test}
	The mirror mark test was independently invented for chimpanzees \cite{gallup1970chimpanzees} and infants \cite{amsterdam1972mirror}. Bard et al.~\cite{bard2006self} provide an excellent overview of the details of the test in the different traditions. Comparative studies (e.g., \cite{gallup1970chimpanzees,anderson2015mirror,deWaal2019fish}) have asked whether chimpanzees, orangutans, elephants, magpies, etc. as a species possess a self-concept; developmental studies (e.g., \cite{amsterdam1972mirror,anderson1984development}) have been concerned with individual differences and developmental milestones. In both traditions, the mirror mark test is a gold standard, an objective assessment, appropriate for nonverbal or preverbal organisms, relying on objective target behavior: reference to the mark on the face, after discovering the mark by looking in the mirror \cite{bard2006self}. However, there are differences in the mark application (infants: spot of rouge applied covertly by mother and placed in a single location alongside nose; chimpanzees: alcohol-soluble dye applied to unconscious animal in two non-visible locations) and testing procedures (e.g., infants: mother scaffolds infant's response). The biggest difference among experiments is probably the interpretation of the responses and judging whether the test has been passed (reference to the mark). According to Mitchell~\cite{mitchell1993mental}, ``the standard evidence of `mirror-self-recognition' is an organism's responding, when placed before a mirror, to a mark on its forehead or other area of the body which is not discernible without the use of the mirror''\cite{amsterdam1972mirror,gallup1970chimpanzees}. However, the detailed assessment criteria differ, as summarized in Table~\ref{table:msr_assessment} (from \cite{bard2006self}). For the purposes of this article, in which we seek explanations for the most low-level, sensorimotor, aspects of MSR, it is the ``touch spot of rouge/mark'' that will be our focus. 
	
	\begin{table}[hbtp!]
	\centering
	%\makebox[1.0\columnwidth]{
\begin{tabular}{ll}
\toprule
\multicolumn{1}{c}{Amsterdam (1972)} & \multicolumn{1}{c}{Gallup (1970)}    \\
\midrule
\multicolumn{2}{c}{Assessment criteria}                                     \\
Recognition of mirror image          & Combination of changing behaviors   \\
 \quad Touch spot of rouge                  & \quad Decrease social responses            \\
 \quad Turns head and observes nose         & \quad Increase self-directed responses     \\
 \quad Labels self (verbal)                 & Touches to mark                      \\
 \quad Points to self                       & \quad Confirms self-directed touches       \\
                                            & \quad More when mirror present than absent
\end{tabular}
%} % end box
\caption{Assessment criteria for mirror self-recognition from \cite{bard2006self}.}
\label{table:msr_assessment}
\end{table}

	\subsection{Mechanisms of mirror self-recognition}
	\label{sec:MSR_mechanisms}
\begin{figure}[htbp!]
	 \includegraphics[width=.42\textwidth]{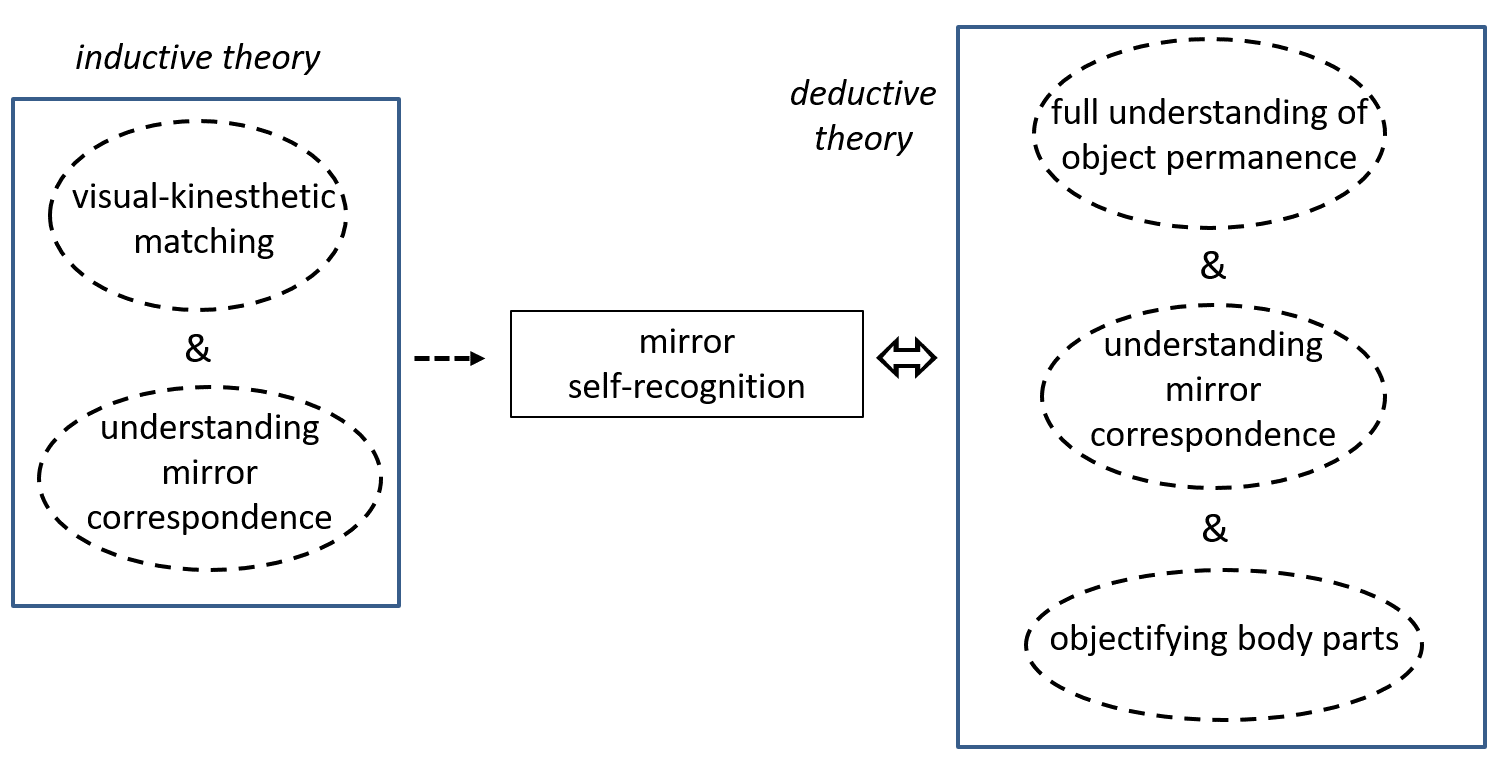} 
	 \label{fig:msr_mitchell}
	 \caption{Inductive and deductive theory of mirror self-recognition according to Mitchell~\cite{mitchell1993mental}. The two theories are enclosed in rectangular blocks, with individual capacities enabling MSR in dashed ellipsoids. The arrow from inductive theory to MSR is dashed as the capacities are likely but not sufficient to enable MSR. The deductive theory provides necessary and sufficient conditions and hence the arrow denoting equivalence.} 
	\end{figure}

	What does success in MSR entail? Mitchell~\cite{mitchell1993mental} proposed two theories, or mental models, which are schematically illustrated in Fig.~\ref{fig:msr_mitchell}: the ``inductive'' and the ``deductive'' theory. The ``inductive theory'' (in agreement with the observations of~\cite{guillaume1971imitation}) presumes that subjects that are (1) capable of visual-kinesthetic matching and that (2) understand mirror correspondence are likely to pass the mirror test. The ``deductive theory'' \cite{mitchell1993mental} should be stronger: if (1) full understanding of object permanence, (2) understanding mirror correspondence, and (3) objectifying body parts is in place, it constitutes a necessary and sufficient condition for MSR. For evidence supporting the theories, the reader is referred to \cite{mitchell1993mental}.

	Let us first look at the inductive theory in detail. Kinesthesia means movement sense but is sometimes equated with proprioception: the subject is aware of where its body is in space based on somatosensory afference. Visual-kinesthetic matching means that the subject can map this information onto a visual image: how such a body configuration would look like. This is necessary for imitation, which is why imitation capabilities are monitored in relation to the likelihood of passing the mirror test. Specifically for MSR, kinesthetic-visual ``self-matching'' is needed. Mitchell~\cite{mitchell1993mental} is not completely clear whether this capacity is primarily spatial (comparing static kinesthetic and visual body configuration images) or temporal (both ``images'' moving in synchrony), but it seems that emphasis is on the former. Adding understanding mirror correspondence---that mirrors reflect accurate and contingent images of objects in front of them---allows the subject to add the mapping needed between the visual body image she constructed and the mirror reflection.
	
	The deductive theory lists three conditions that are necessary and sufficient for MSR. The first prerequisite is full understanding of object permanence, corresponding to Piagetian stage 6 of this capacity \cite{Piaget1954}, which ``presupposes that an organism has memory and mental representation, recognizes similarity between similar objects, recognizes that its body is a continuous object localized and extended in space (and therefore \textit{represents its body as such} in this way), ..., and has some primitive deductive abilities.'' \cite{mitchell1993mental} The second prerequisite, understanding mirror correspondence, has been discussed above. By objectifying body parts, the third capacity, ``is meant both (1) that the organism recognizes the similarity between any particular body-part of-its own and the same body-part of another (a recognition presupposed by understanding object permanence), and (2) that the organism recognizes a body-part, and recognizes it as part of the body, even when the body-part is decontextualized---that is, separated from the body.'' \cite{mitchell1993mental} Then, according to Mitchell, ``the organism perceives \textit{x} (its hand) as an object which is distinct yet continuous with \textit{y} (its body), and knows that mirrors reflect objects accurately and contingently; if \textit{x} is an object distinct yet continuous with \textit{y}, and if mirrors reflect  objects accurately and contingently, then if a mirror reflects \textit{x}, it must simultaneously reflect \textit{y}; the organism knows that the mirror reflects \textit{x}; therefore the organism knows that the mirror reflects \textit{y} and thus recognizes the mirror-image as an image of its body.'' \cite{mitchell1993mental}
	
    Passing the test without ``cheating'' assumes, first, that the subject identifies herself in the mirror (``it's me in the mirror'' box). That is, if the subject thinks it is her conspecific in the mirror with a mark on the forehead and then goes on to explore also her forehead, it should not count as passing the test and should be controlled for. Gallup~\cite{gallup1982self} postulates ``an essential cognitive capacity for processing mirrored information about the self''. Animals that possess this capacity or that ``are self-aware'' can succeed. Alas, such an explanation does not bring us closer to understanding the mechanisms. Anderson~\cite{anderson1984development} assumes that recognition of one's body-image in a mirror results from ``a mental representation of self onto which ... perception of the [mirror] reflection is mapped'' (cited from \cite{mitchell1993mental}).\footnote{Mitchell~\cite{mitchell1993mental} discusses the ``chicken-and-egg problem'' of acquiring this self-image---prior recognition in the mirror may be necessary to learn it---and concludes that there are ``three possibilities: (1) a visually based, incomplete self-image of the part of the organism it can see, (2) a non-visual self-image, (3) or a mixture of these images.''} 
	
	\begin{figure*}[ht!]
		\centering
		\includegraphics[width=.95\textwidth]{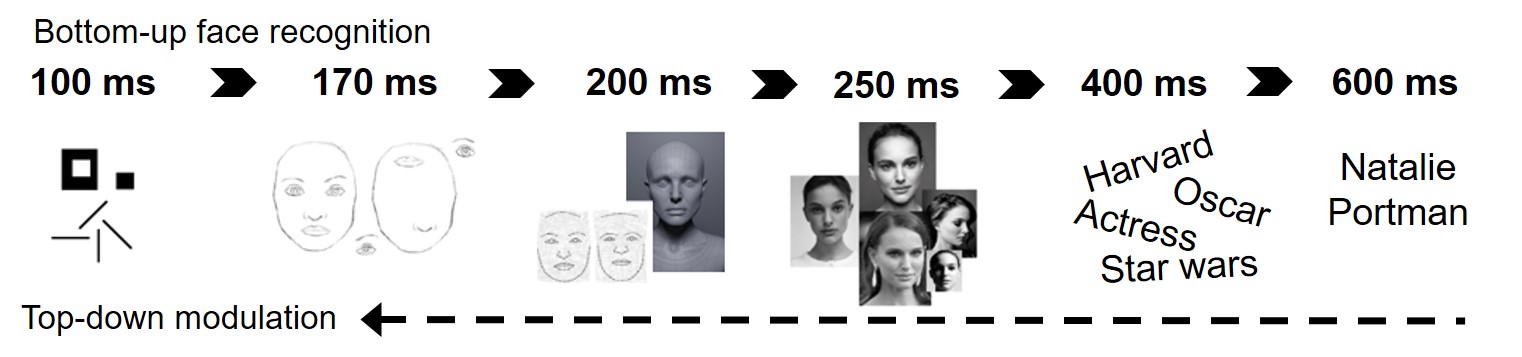}
		\caption{Human face self-recognition process.}
		\label{fig:face_recog_humans}
	\end{figure*}
	
	However, a ``self-image'' is not the only way of getting at ``it's me in the mirror''. ``Temporal contingency (image moves as I move)'' may presumably be more effective---bypassing the problem of visual matching dependent on image translations, rotations, size, clothes in case of infants etc. This is an instance of the general question of body ownership vs. agency \cite{tsakiris2010having}: which of them, or perhaps their combination, is relevant to pass the ``it's me in the mirror'' in the MSR context?
	Bigelow~\cite{bigelow1981correspondence} provides evidence for the temporal contingency cue in a study where movement was used as a cue to self-recognition in children by presenting them with movies of their own or other infants' photographs in or out of sync with their movements.
	Thus, an image of how one's whole body looks from the outside may not be necessary: the subject may do away with identifying it is her through temporal contingency and then an image of her face may suffice. However, in practice, temporal and visual contingency are  intertwined during the MSR test.
	
	To succeed in the test, the subject needs to display one of the behavioral responses listed in Table~\ref{table:msr_assessment}. To display a response, the subject needs ``motivation'', which  has also generated some controversy in the field: the responses should be spontaneous and engineering them in species that do not normally pass the test through reinforcement---monkeys in particular--- has been criticised \cite{anderson2015mirror,deWaal2019fish}. 
	The response we focus on is ``touch spot of rouge'' / ``touch to the mark''.
	There are several ways of reaching for the mark. One key distinction is whether the reaching is ``feedforward'' or ``feedback''. In the study of reaching development, the ``visually guided reaching'' hypothesis---infants need to look at their hands and the object alternately in order to progressively steer the hand closer to the object location---has been replaced by ``visually-elicited'' reaching, whereby the infant looks at the target and continues to do so during the reaching (Corbetta et al.~\cite{Corbetta2014} provide an overview and additional evidence). If the latter strategy was employed to reach for the mark, first, some form of ``remapping of stimulus location in the mirror to an ego-centric frame'' is necessary. Heed et al.~\cite{Heed2015} review the process of ``tactile remapping'': how a stimulus on skin may be transformed such that it can be acted upon (looked at, reached for, ...). However, how that is achieved in the case of the mirror remains an open question. Some ``understanding of mirror correspondence'' is probably recruited. However, it seems very unlikely that such understanding will be so mature that the mark could be localized in space with respect to the body frame. Instead, it seems more likely that the mark location is identified with respect to some landmarks on the face (nose, mouth, etc.) encoded in lower hierarchical levels of visual face recognition \cite{Eimer2012}. Then, the rest of the localization process may be similar to tactile remapping.
	
    The role of the tactile modality remains to be further explored. The experiments of Chang et al.~\cite{chang2015mirror} with macaques show that making the association between the visual stimulus in the mirror and tactile sensation (using high-power laser) facilitates later success in the mirror test. Perhaps, this association is needed to ``bootstrap'' the remapping process. This is hypothesized also by de Wall~\cite{deWaal2019fish}: ``it is as if these animals need multimodal stimulation to get there'' (also called ``felt mark test'' in \cite{deWaal2016we}).
		
    An alternative approach to MSR would be to consider the brain as a Bayesian machine that encodes sensory information in a probabilistic manner. In this scheme, self-recognition is achieved by differentiating our body from other entities because they are probabilistically less likely to generate the observed sensory inputs~\cite{apps2014free}. This inevitably requires the ability of encoding priors (by learning) and the availability of generative processes that predict the effects of our body in the world~\cite{friston2010free,lanillos2020robot}. For instance, within this approach, visual congruency may be computed through minimizing the prediction error between the expected appearance and the current visual input at each hierarchical layer---a predictive coding approach \cite{rao1999predictive}.  Under the Free-energy principle \cite{friston2010free,apps2014free}, self-recognition is related to low surprise in the sensory input. The brain is able to correctly predict and track the body effects in the world. 
    
    The behavioral response and its awareness in the MSR is, however, a controversial aspect. Friston  \cite{friston2010action,friston2010free} proposed that the action minimizes the prediction error derived from proprioceptive expectations by means of the classical motor reflex pathway. Assuming that salient regions in the face (e.g., mark) produce a goal driven response to reduce this visual ``error'', we would expect ``visually guided reaching'' as described above. It is still debatable that the behavioral response produced in this scheme will encompass self-awareness as studied in primates. This will require some level of body-ownership and agency using the mirror reflection \cite{zaadnoordijk2019match}.

	\subsection{Face and self-face recognition in the brain}
	\label{sec:face_recog}
	According to the cognitive and functional model of facial recognition in humans \cite{Bruce1986,Gobbini2007}, recognition occurs through a hierarchical process, as depicted in Figure \ref{fig:face_recog_humans}. Initially, facial low-level visual information is analyzed. Then, facial elements (eyes, nose, mouth) are detected \cite{Eimer2012} and the spatial relationship between them are integrated forming a layout \cite{Latinus2006}. Once the physical characteristics of the face have been coded by the visual system, the resulting representation must be compared with the known faces representations stored in long-term memory---at the Facial Recognition Units (FRUs) \cite{Schweinberger2016}. Only when there is high correspondence between the computed and the perceived representation, there is access to the semantic and episodic information related to the person (relationship, profession, etc.), and finally to his/her name \cite{Schweinberger2016}. This last stage happens at the so-called Person Identity Nodes (PINs), which can be activated by different sensory pathways (e.g., auditory by someone's voice).
	
	This general face recognition slightly differs when recognizing our own face. Studies that investigate the temporality of self-face recognition show that self-face differs from general face recognition already at an early stage in visual processing (around 200 ms after the stimulus onset) \cite{Alzueta2019}. At this stage, self-face processing is characterized by a reduced need for attentional resources. This bottom-up attention mechanism facilitates the activation of self-face representation on memory (FRUs) and therefore self-recognition. Surprisingly, once self-face has been recognized, a top-down attentional mechanism comes into play allocating cognitive resources on the brain face-related areas to keep self-face representation in active state \cite{Alzueta2020}. 
	
	Neuroimaging studies also evidence the interplay between bottom-up and top-down attentional control brain areas during self-processing \cite{Sui2017}. The activation of a specific Self-Attention Network supports the theoretical view that attention is the cognitive process more distinctive during self-face recognition. In the case of the mirror test, this increased attentional mechanism would strengthen the relevant signals such as novel visual cues on the face (e.g., mark) as well as boost the access to memory and produce the feeling of awareness.
	
	\subsection{Robot self-recognition}
	
	Several works have addressed self-recognition in robots (works until 2010 reviewed in \cite{hoffmann2010body} and revisited in \cite{lanillos2017enactive} from the enactive point of view). First, we describe works on body self-recognition and second, we summarize works that specifically studied robots in front of a mirror.

	%There can be different forms of self-recognition, the majority of them not requiring a mirror. We will review these first. Second, we will summarize works that specifically studied robots in front of a mirror.}
	
	Two principally different strategies were employed for machine self-recognition. According to the first, the body is the invariant: what is always there. The research of Yoshikawa and colleagues (e.g., \cite{yoshikawa2004my}) capitalizes on this property, acquiring a model of ``how my body looks like''. In \cite{laflaquiere2019self,diez2019sensorimotor}, the robot learns the appearance of its hand and arm using deep learning techniques. The second strategy takes a largely opposite approach: my body is what moves, and, importantly, what I can control. Fitzpatrick and Metta  \cite{Fitzpatrick2002} exploit the correlation between the visual input (optic flow) and the motor signal; Natale et al. \cite{Natale2007} improve the robustness of this procedure by using periodic hand motions. Then, the robot's hand could be segmented by selecting among the pixels that moved periodically only those whose period matched that of the wrist joints. 
	
	Bayesian and connectionist approaches have been proposed to capture this sensorimotor correlation for self-recognition. Tani, in \cite{tani1998interpretation}, presents self-recognition from the dynamical systems perspective using artificial neural networks. Gold and Scassellati \cite{gold2009using} employ probabilistic reasoning about possible causes of the movement, calculating the likelihoods of dynamic Bayesian models. A similar approach was proposed in \cite{lanillos2016self,lanillos2016yielding}, where the notion of body control was extended to sensorimotor contingencies: ``this  is  my  arm  not  only  because  I  am  sending  the command to move it but also because I sense the consequences of moving it''. All these exploited the spatio-temporal contingency, related to the sense of agency. Pitti et al. \cite{pitti2009contingency} studied temporal contingency perception and agency measure using spiking neural networks. Gain-field networks were employed to simultaneously learn reaching and body ``self-perception'' in \cite{abrossimoff2018visual}.
    
    Specifically, mirror self-recognition has also been studied in robots. Steels and Spranger \cite{steels2008robot} explored this situation from the perspective of language development. A Nao robot was engineered to pass the mirror test using logical reasoning in \cite{bringsjord2015real}. Hart \cite{hart2014robot} employed state-of-the-art techniques in computer vision, epipolar geometry, and kinematics. Fuke et al. \cite{Fuke2007} proposed a model in which nonvisible body parts---the robot's face---can be incorporated into the body representation. This was done via learning a Jacobian from the motor (joint) space to the visual space. A neural network with Hebbian learning between the visual, tactile, and proprioceptive spaces was used. Integrating the velocities, position in visual space can be estimated for nonvisible parts as well. Then, while the robot was touching its face with the arm, the position in the visual modality could be estimated and matched with the touch modality, learning a cross-modal map. 
    
    Finally, Lanillos et al. \cite{lanillos2020robot} recently proposed an active inference (i.e. free-energy principle) approach to MSR where the robot learned to predict the sensory consequences of its actions using neural networks in front of the mirror and achieved self-recognition by means of evidence (absence of prediction error) accumulation.
    
    In most of the works cited above, self-recognition in general or MSR in particular was largely engineered. In this work, we present a process model that is more tightly grounded in the psychological literature on MSR. Furthermore, we present an embodied computational model on a humanoid robot, in which the novelty detection is currently our main contribution.
    
	\section{PROCESS MODEL OF MIRROR SELF-RECOGNITION}
	\label{sec:mechanisms}
\begin{figure}[hptb!]
		\centering
		\includegraphics[width=.45\textwidth]{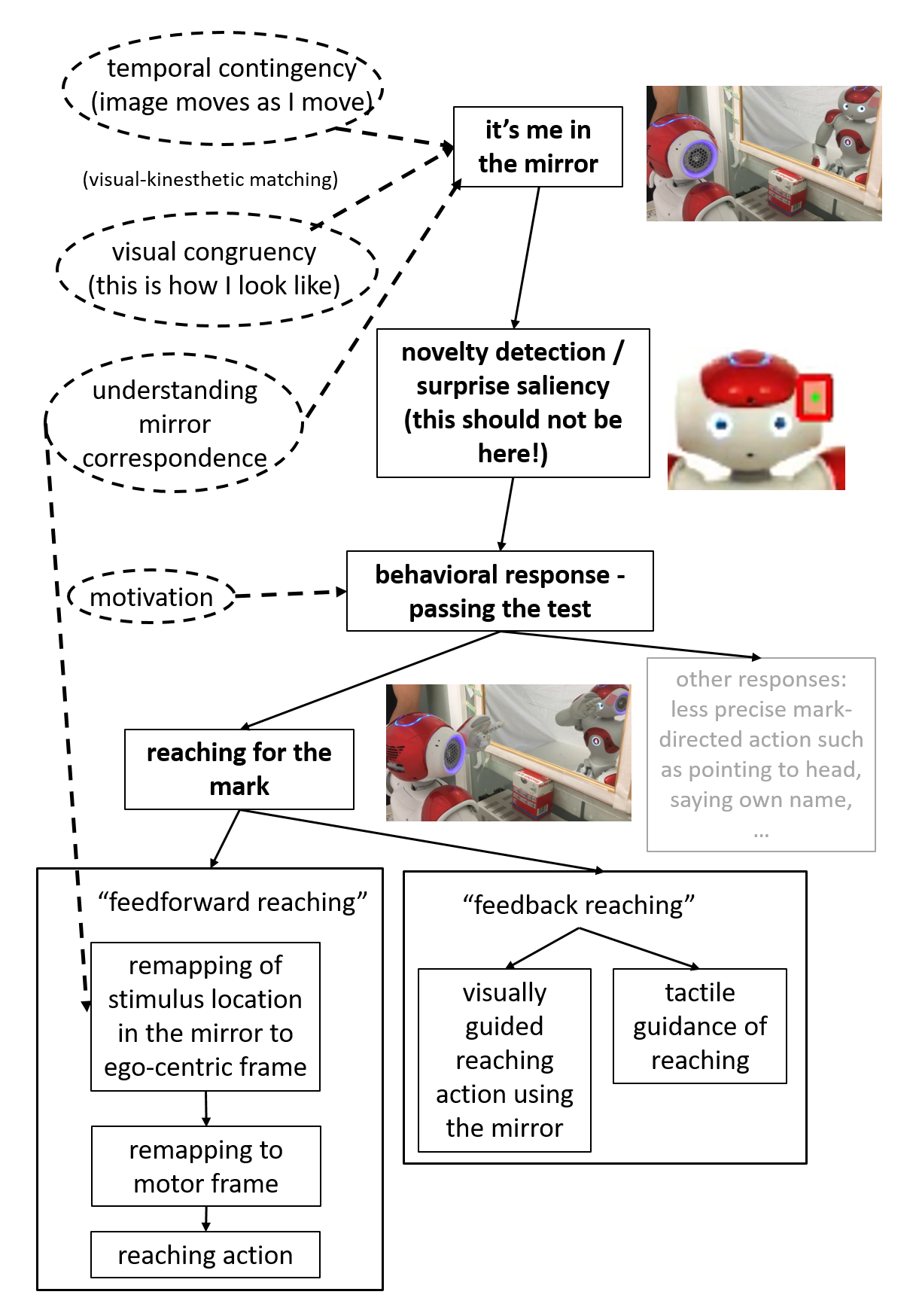}
		\caption{Process model of mirror self-recognition. The main pipeline is illustrated using blocks with text in bold. From top to bottom: self-recognition in the mirror, identifying the mark,  reaching movement. The control of the reaching movement can be feedforward (left) or feedback (right). Circles with dashed lines illustrate ``prerequisites'' for individual blocks.}
		\label{fig:msr_mechanism_schematics}
	\end{figure}

    Mitchell's theories (\cite{mitchell1993mental} and Section~\ref{sec:MSR_mechanisms}) suggest possible modules or components that may be needed for MSR. However, they are still quite abstract high-level capacities and their specific role in the \textit{process of passing the mirror test} is unclear. Therefore, we propose instead a \textit{process model} (Fig.~\ref{fig:msr_mechanism_schematics}) of going through MSR. We do employ ``visual-kinesthetic matching'' and ``understanding mirror correspondence'' blocks in our model, while we leave ``fully understanding object permanence'' and ``objectifying body parts'' aside---these are very complex capacities that would be far from trivial to implement. Instead, we hope that their explicit instantiation is not necessary for MSR. Possibly, behavior that can be interpreted as such in this context may emerge in our model. Our proposed mechanistic account is shown in Fig.~\ref{fig:msr_mechanism_schematics}. 
    
    \subsection{It's me in the mirror}
  
	The first ``block'' in the process model of MSR is ``it's me in the mirror'', which likely arises from some visual-kinesthetic matching as discussed in Section~\ref{sec:MSR_mechanisms}, and takes the form of ``visual congruency'' (``this is how I look like'') stressed by \cite{mitchell1993mental}, or ``temporal contingency'', or their combination.   
	Additionally, ``understanding mirror correspondence'' likely also contributes to the ``it's me'' test. It seems plausible to think that it would be relatively more important for the ``visual congruency'' cue, whereby the image of the self needs to be matched against its specular reflection.

	\subsection{Novelty detection / surprise saliency}
	The ``novelty detection / surprise saliency'' is the module responsible for recognizing one's own face (Section~\ref{sec:face_recog}) and detecting the mark as an object that does not belong there. It is this part that our computational model on the robot will specifically address (Section~\ref{sec:model})---constituting the main technical contribution in this work.
	
	\subsection{Reaching for the mark}
	To succeed in the test, the subject needs to display one of the behavioral responses listed in Table~\ref{table:msr_assessment}. Our focus will be the ``reaching for the mark'' and we will leave the ``other responses'' aside, as these do not easily lend themselves to a mechanistic decomposition. To display a response, the subject needs ``motivation''. However, the responses should be spontaneous and not engineered  through external rewards. From the perspective of our process model, the problem with such reinforcement is that the subject may learn to pass the test while side-stepping the ``it's me in the mirror'' and ``novelty detection'' blocks.  
	
	How the ``reaching for the mark'' is performed remains an important open question. The subjects passing MSR should be capable of ``feedforward reaching''. If this strategy is employed to reach for the mark, first, some form of ``remapping of stimulus location in the mirror to ego-centric frame'' is necessary. It is not clear how this is done in this case. ``Understanding mirror correspondence'' will facilitate the localization. However, it seems unlikely that such understanding will be so mature that the mark location could be remapped into, say, the body frame through a combination of stereo vision and mirror projection. Instead, it seems more likely that the mark location is identified with respect to some landmarks on the face (nose, mouth, etc.).  Then, the rest of the localization process may be similar to tactile remapping. Next, the target location may be transformed to ``motor coordinates'' and finally executed.

    Alternatively, it could be that in this unusual situation, the subjects would employ a ``visually guided reaching action using a mirror''. Furthermore, in case the initial reach is not accurate, ``tactile  guidance of reaching'' is also a possibility. All these options can in principle be realized in a robot. However, more information from experiments with infants and animals is needed to provide the right constraints for the model.

	\section{MIRROR RECOGNITION IN A ROBOT}
	\label{sec:model}
	
\begin{figure}[hptb!]
		\centering
		\includegraphics[width=.45\textwidth]{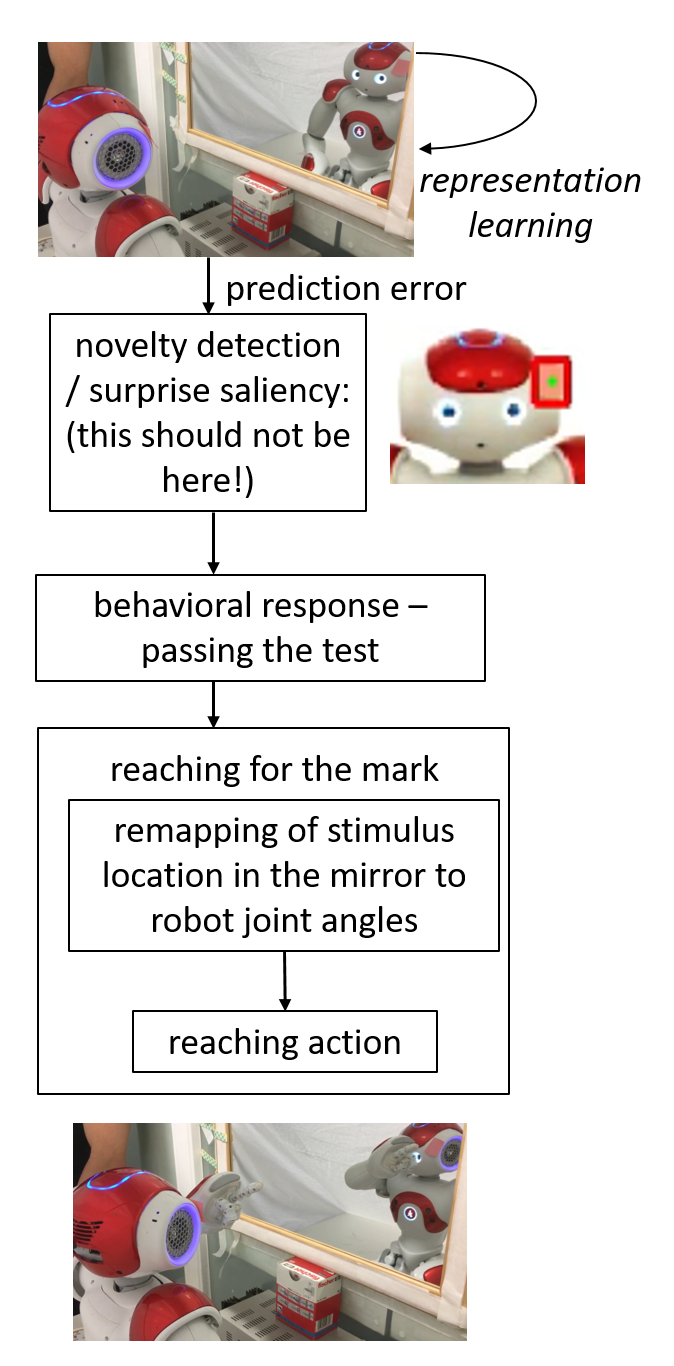}
		\caption{Schematics of mirror self-recognition implemented in this work on the robot.}
		\label{fig:workflow_robot}
	\end{figure}

	In this section, we present the architecture that we implemented on a humanoid robot -- Fig. \ref{fig:workflow_robot}. Compared to Fig.~\ref{fig:msr_mechanism_schematics}, it is in many ways simplified. The novelty detection / surprise saliency was modeled in detail, inspired by the the predictive coding hypothesis. In order to simplify the system, we will assume that the robot is able to identify itself in the mirror, triggering the top-down modulation and attentional capture. That is, the high-level hierarchies of the face recognition process described in Fig. \ref{fig:face_recog_humans} are not addressed. Besides, the reaching for the mark has been grossly simplified at the moment.
	
	\subsection{Learning the appearance representation}
	\label{sec:representationlearning}
\begin{figure}[hbtp!]
		\centering
		\subfigure[Autoencoder]{      \includegraphics[width=.32\textwidth]{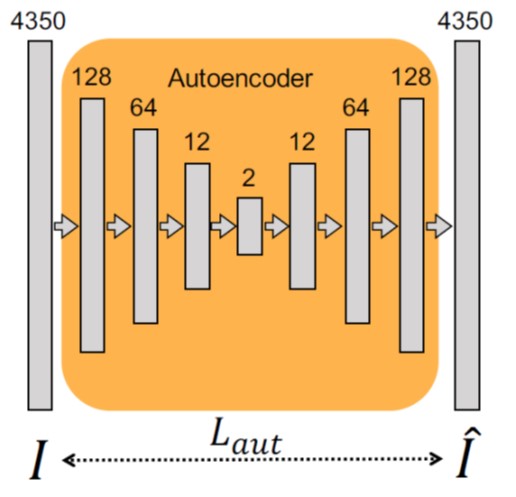} \label{autoencoder}}
		\subfigure[Decoder]{      \includegraphics[width=.32\textwidth]{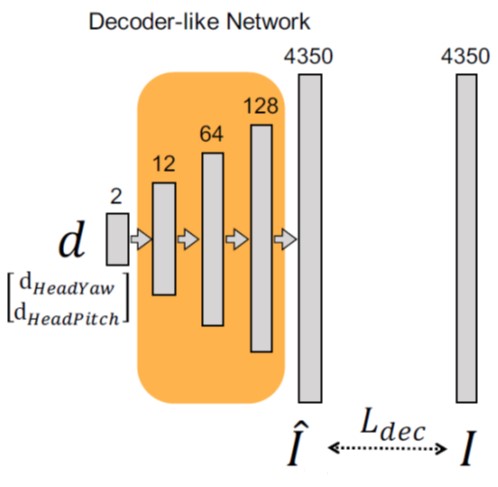} \label{decoder}}
		
		\caption{Generative model learning. Two convolutional neural networks were tested: (a) autodecoder and (b) decoder with joint angles as the input. The number of nodes for each layer used is detailed.}
		\label{fig:generativemodel}
	\end{figure}

	We assume that the robot has some kind of self-distinction abilities using non-appearance cues \cite{lanillos2016yielding,lanillos2020robot}. Therefore, it can learn the face representation in front of the mirror through semi-supervised learning. This allows the robot to imagine or predict its face appearance in the visual space. We studied two different network architectures described below and depicted in Fig. \ref{fig:generativemodel}.
	
	\subsubsection{Self-supervised autoencoder}
	
	The first architecture, depicted in Figure \ref{autoencoder} is known as an autoencoder \cite{ballard1987modular}. We used this artificial neural network (ANN) to learn the high level visual features of the face, analogously to the visual face recognition in humans. The network also learns the generative process or ``visual-kinesthetic'' forward model predictor. Given the input image $I$ and the predicted image $\hat{I}$, the network was trained using the MSE reconstruction loss: $L = \frac{1}{N} \sum (I-\hat{I})^2$, where $N$ is the number of pixels in the image. We used $\tanh$ as the activation function in each layer except the output layer, where the \texttt{Sigmoid} function was selected to obtain the desired pixel output in the range of $(0, 1)$.
		
	\subsubsection{Supervised decoder-like neural network}
	Inspired by the decoder part of the autoencoder architecture, we built a decoder-like neural network, as shown in Figure \ref{decoder}, where the latent space was substituted by the joint encoders of the robot head. Similar architectures have been used for learning the visual forward model \cite{laflaquiere2019self,sancaktar2020end}.
	We denote $\bm{d} = [d_{HeadYaw}, d_{HeadPitch}]$ the head motor state as input vector to the network. Analogous to the autoencoder approach, $\tanh$ was the activation function in each internal layer and \texttt{Sigmoid} as the output layer. The predicted image $\hat{I}$ is compared with the original image $I$ corresponding to the input motor state $\bm{d}$. We also used the MSE between $\hat{I}$ and $I$ as training loss. Here, the motor state generates the corresponding image.
	
	\subsection{Visual novelty detection using generative models and the prediction error}
	\label{sec:novelty}
	Once the appearance representation is learned, the robot can directly use the generative model to discover novel visual events, such as a colored mark placed in the face. Surprising events will have high prediction error. Therefore, we computed the image saliency by subtracting the predicted visual input and the current observation ($I-\hat{I}$). In practice, once the generative model is trained, the robot can predict its visual appearance depending on the head angles and compute the visual prediction error. However, simply computing this difference would lead to inaccurate distinction between highly variable regions (e.g. the eyes, mouth, light reflections, etc.) from real novel visual events. We need to take into account the variance associated with each pixel information. Hence, we computed the distribution that encodes mean prediction error for each pixel $\mu$ and its variance $\sigma^2$ as follows:
	
	\begin{equation}
	\mu = \frac{1}{N_t}\sum_{i=1}^{N_t} |I_i - \hat{I}_i|
	\end{equation}
	\begin{equation}
	\sigma^2 = \frac{1}{N_t - 1}\sum_{i=1}^{N_t} (|I_i - \hat{I}_i| - \mu)^2
	\end{equation}
	where $N_t$ denotes the number of collected images and $\hat{I}_i$ is the $i^{th}$ predicted face generated by the network. Finally, the saliency map $I_s$ was computed by means of the Mahalanobis distance ($D_M$) :
	\begin{equation}
	I_s = D_M(I,\hat{I}) = (|I-\hat{I}| - \mu)/\sigma^2
	\label{eq:dm}
	\end{equation}
	where all operations are pixel-wise, $I, \hat{I}$ are current specular image and the predicted image, respectively. Note that we are assuming that there is no correlation within pixels and thus, the covariance matrix $\Sigma$ is diagonal and it is defined by the $\sigma^2$ vector.
		
	\subsection{Reaching for the mark}
	\label{sec:reaching_for_mark}
	In the current model, the reaching behavior has been greatly simplified and engineered. It could be said the our solution corresponds to the ``feedforward reaching'' strategy of Fig. \ref{fig:msr_mechanism_schematics}; however, no remapping from the image in the mirror to an egocentric reference frame is performed. Instead, the robot head and arm were manually driven into configurations in which they reach for the mark at different locations on the face. The mapping from the detected novelty region in the specular image to the robot joint angles is directly learned (Fig. \ref{fig:workflow_robot}).   
     To this end, we used a feedforward neural network to map the centroid position of the mark in the visual space $c = (i, j)$ to joint states $q$ (e.g., head yaw, shoulder pitch). The ANN had three layers: input layer with 2 neurons for the mark location in the mirror, one hidden layer (10 neurons) and output layer with the number of neurons matching the number of joint states. We used the Sigmoid as the activation function in each layer.
	
	Once this mapping, $(i,j)\rightarrow \mathbf{q}$, is learned, the points in the image frame in pixels have a direct translation to the joint angles space. 
	The target joint angles are eventually sent to the joint position control of the Nao robot and the movement is achieved via local PID motor controllers.
	
	\section{EXPERIMENTS AND RESULTS}
	\label{sec:results}
	Here we describe the experimental setup and the implementation details of the model deployed. We  quantitatively evaluated the face representation learning and novelty detection system and qualitatively the robot behavior with two different Nao robots. A supplementary video showing the system in action can be found here: \url{https://youtu.be/lbdAyOPkIIM}.
	
	\subsection{Experimental setup}
\begin{figure*}[hbtp!]
		\centering
		\includegraphics[width=.8\textwidth]{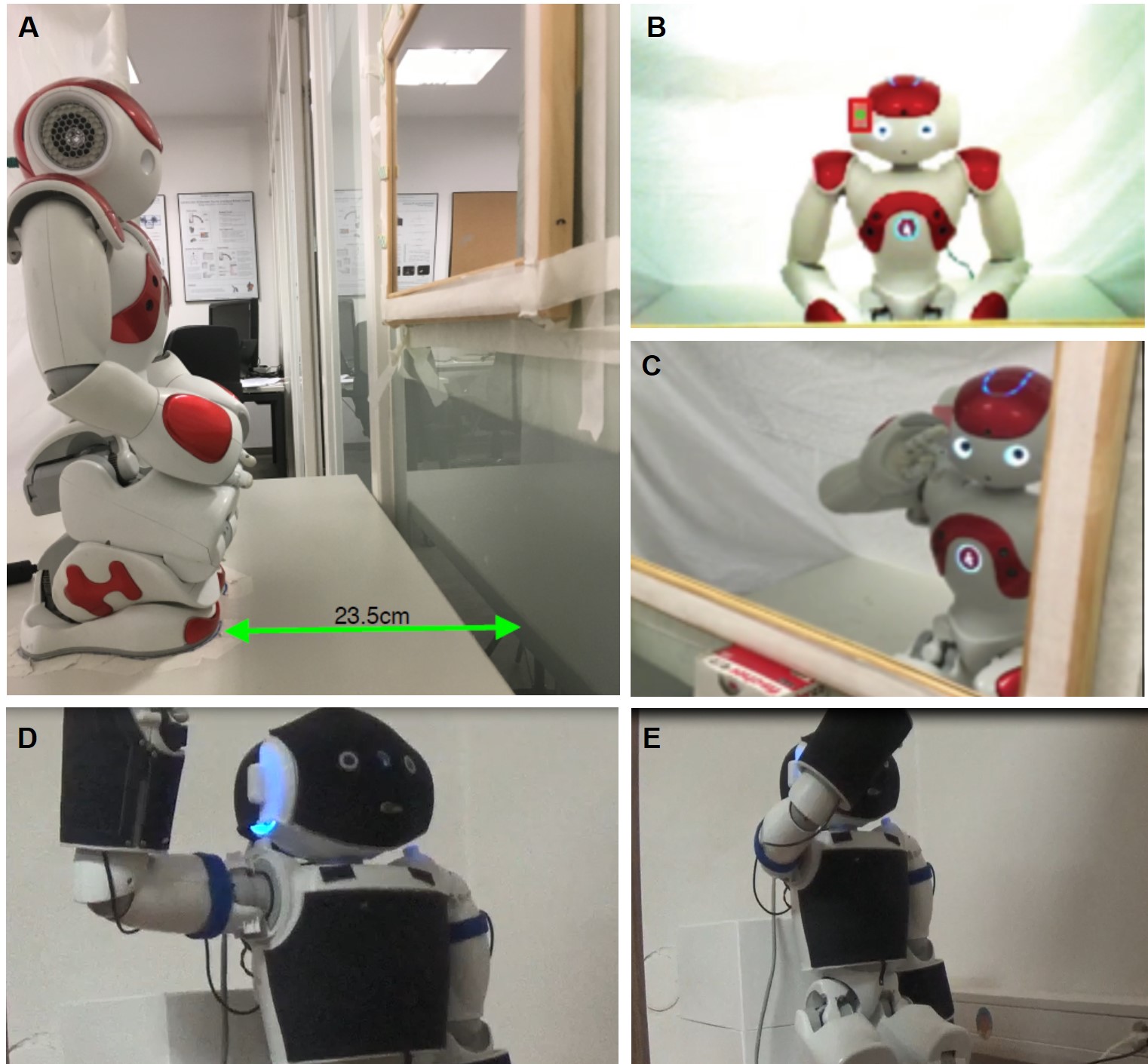}
		\caption{Experimental setup. (A) A Nao robot was placed in front of the mirror for all trials. (B) Mark detection using the generative model prediction error approach. (C) Behavioral response towards the mark. (D) Mirror reflection of the second Nao with electronic skin on the face. (E) Reaching behavior in the second Nao.}  \label{settingup}
	\end{figure*}

	The experimental setup is shown in Figure \ref{settingup}. We tested the approach in two Nao robots with different appearance in both simulation and using the real platforms. We placed the robot facing a mirror at a fixed distance, leaving the face and the torso visible for both training and test trials.
		
	\subsection{Training and evaluation}
	
	\subsubsection{Learning the face representation}
	To evaluate the performance of the representation learning, we collected 1300 mirror reflection images in the Gazebo simulator and transformed into grayscale. Nao camera resolution is $1280\times960$. During training, head yaw and pitch were randomly sampled in a range of $-5^{\circ}$ to $5^{\circ}$. The head region of the robot was cropped out by using the \textit{OpenCV} function \texttt{matchTemplate} using an example robot head as a template. 
	
	We randomly selected $80\%$ of the collected images as our training set, while the rest ($20\%$) were assigned as test set. For the test dataset, we modified the images by synthetically adding a mark (e.g., a rectangle of $14\times 14$ pixels) with random color (from a set of predefined colors) placed in a random position in each test image. The selection of the mark location and color was uniformly distributed.
	
	We used the ADAM optimizer \cite{kingma2014adam} to train both ANNs described in Sec. \ref{sec:representationlearning} and all hyperparameters were fixed equally for both models. Specifically, the number of training epochs was set to $50$. A random mini-batch with the size of $N=128$ was used to train the models for one iteration in every epoch. The initial learning rate was set to $0.005$, and it decreased after every 10 epochs by a decay factor $0.5$.

	\begin{figure}[htbp!]
		\subfigure[Training and test loss]{      \includegraphics[width=.42\textwidth]{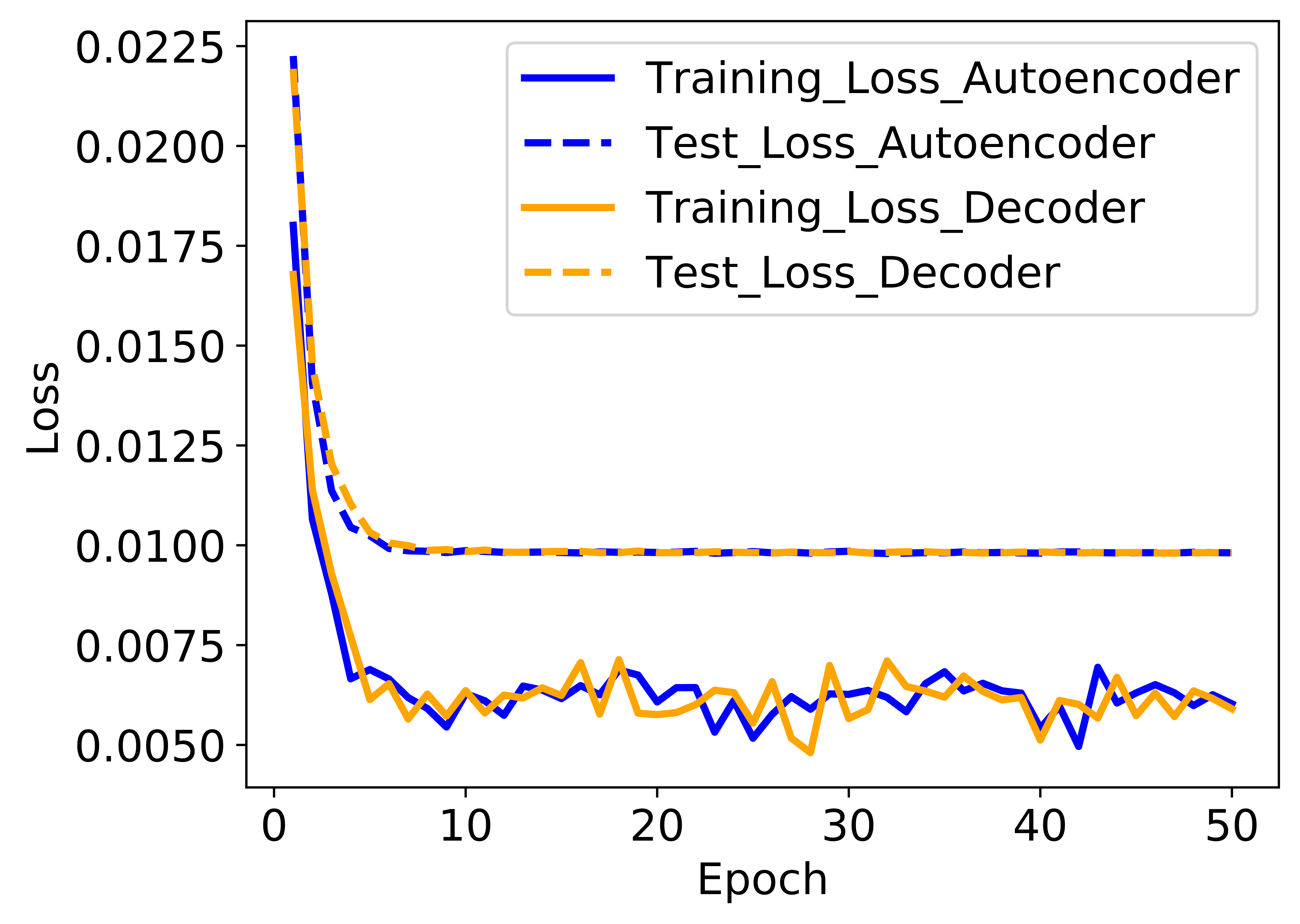} \label{loss}}
		\subfigure[Autoencoder latent representation]{      \includegraphics[width=.47\textwidth]{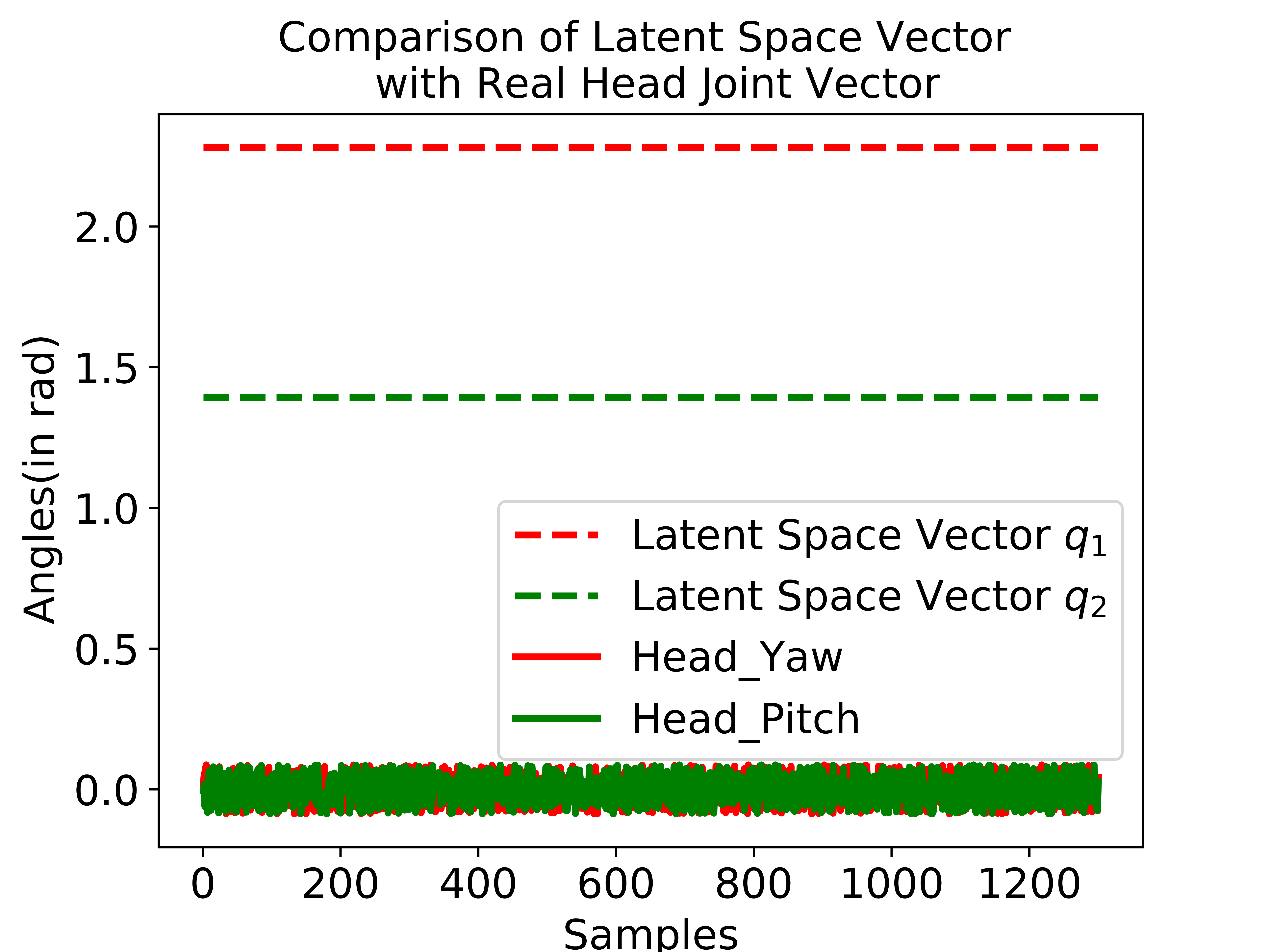} \label{code}}
		\caption{Learning face representation -- training and testing. (a) Autoencoder and decoder-like model loss on training and test dataset during the training process. (b) Autoencoder latent space vs. head joint state vectors. The encoded representation did not match the joint states.}
	\end{figure}
		
	Figure \ref{loss} shows the training loss curves for both training and test sets during the training process. Both models achieved similar reconstruction accuracy, converging to a MSE error of 0.01 on the test set. The autoencoder latent representation learned did not match the head orientation of the robot (i.e. head joint motor states $d_{HeadYaw}$ and $d_{HeadPitch}$), as shown in Fig. \ref{code}.
		
	\subsubsection{Novelty detection}
	
	After we trained the models, we compared the performance of the two different ANN architectures by evaluating the accuracy for detecting the mark in the face. First, the saliency image $I_s$ was binarized by selecting the pixels where the $D_M$ (Eq. \ref{eq:dm}) was greater than $1.8\%$ of the maximal pixel value in greyscale. Afterwards, areas which contained at least 30 consecutive saliency pixels were taken as relevant regions. The most salient area (in a winner-take-all manner) was selected as the output (i.e. mark) from the algorithm for evaluation purposes.
	
\begin{figure}[htbp]
		\centering
		\subfigure[Evaluation Measure]{
			\begin{minipage}[t]{0.6\linewidth}
				\includegraphics[width=1\textwidth]{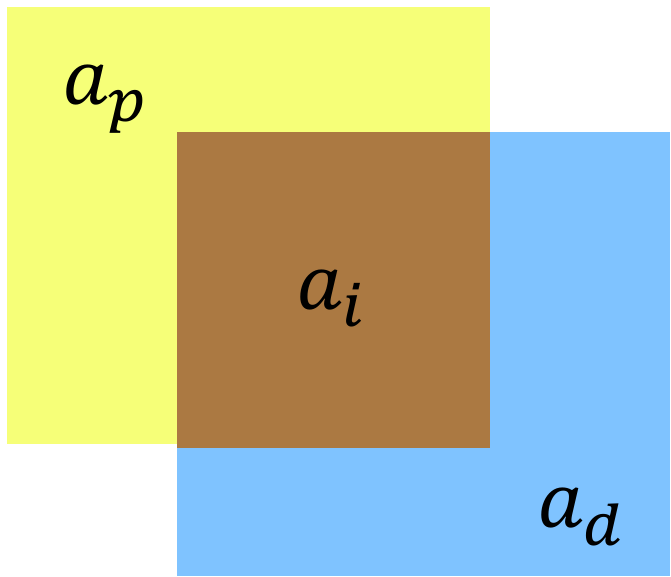}
			\end{minipage}
		\label{measure}
		}	
		
		\subfigure[Accuracy comparison]{
			\includegraphics[width=0.48\textwidth]{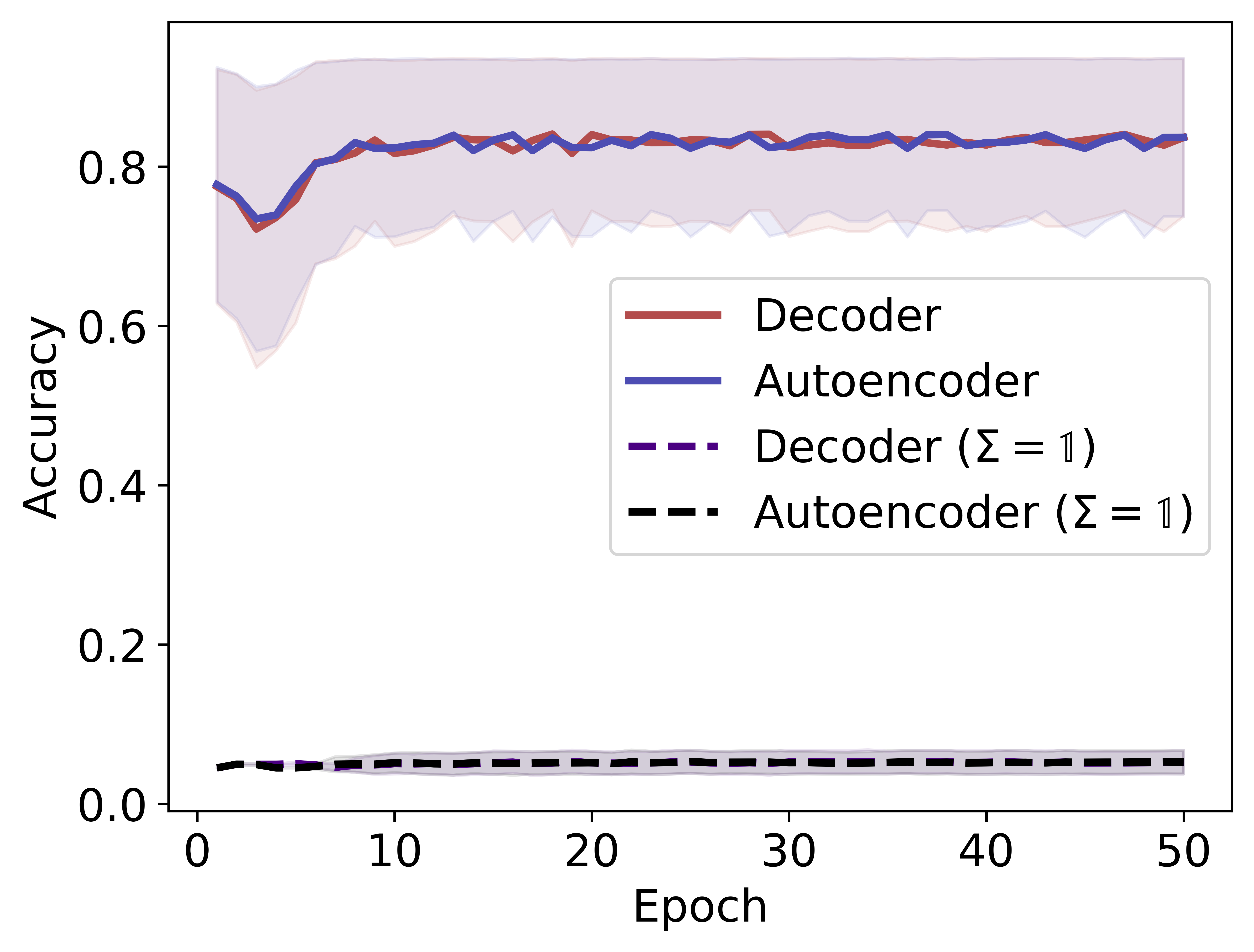}
		\label{accuracy}
		}		
		\caption{Evaluation of the novelty detector. (a) Definition of the overlapping areas of the visual mark and the detected salient region: $a_p$ denotes the area of the mark minus the intersected area $a_i$; $a_d$ denotes the detected area except the intersected area (i.e., false positives). (b) Accuracy comparison between the two ANN architectures with and without prediction error weighted by the variance ($\Sigma = diag(1)$).}
		
	\end{figure}

	To evaluate the novelty detection performance, we defined the precision measure metric~\cite{lanillos2016yielding}: intersected area divided by the sum of intersected and detected area ($a_i / (a_i + a_d)$) -- see Figure \ref{measure}. Values close to one mean that both regions overlap.
	
	Figure \ref{accuracy} shows the accuracy comparison for the two generative models. The measure was calculated at every epoch by averaging its value over all the test set during the training process. Both architectures obtained similar results and converged to a steady state in less than 20 epochs. Results indicate that the prediction error variance was critical and more important than the prediction error in order to properly segment the mark from the face. In particular, when setting the prediction error variance to 1, many regions of the face become salient and the mark was not properly segmented returning low values of our metric.

	\begin{figure*}[t]
		\centering
		
		\subfigure[hexagonal mark]{			
				\includegraphics[width=0.15\textwidth]{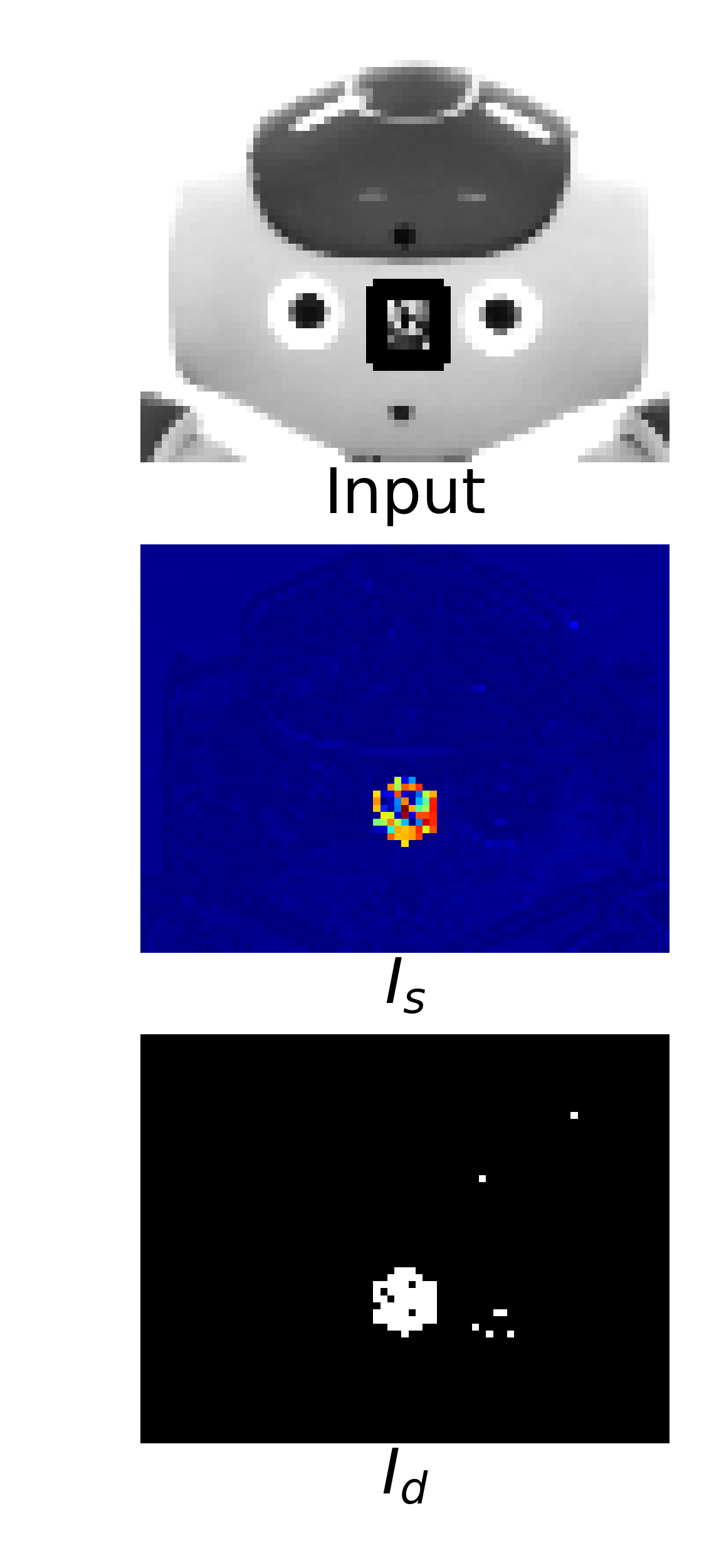}
		}
		\subfigure[big post-it]{			
			\includegraphics[width=0.15\textwidth]{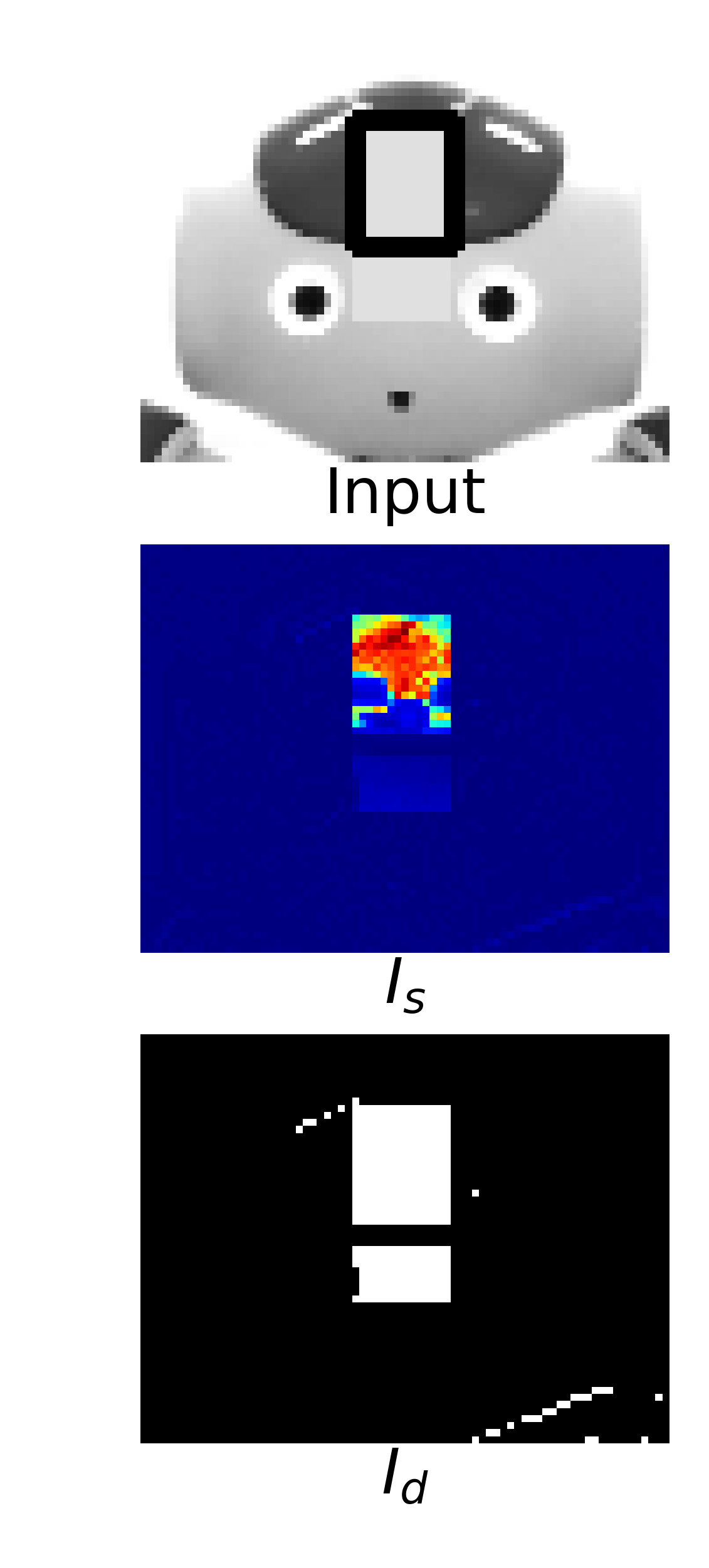}
		} 
		\subfigure[yellow post-it]{			
			\includegraphics[width=0.15\textwidth]{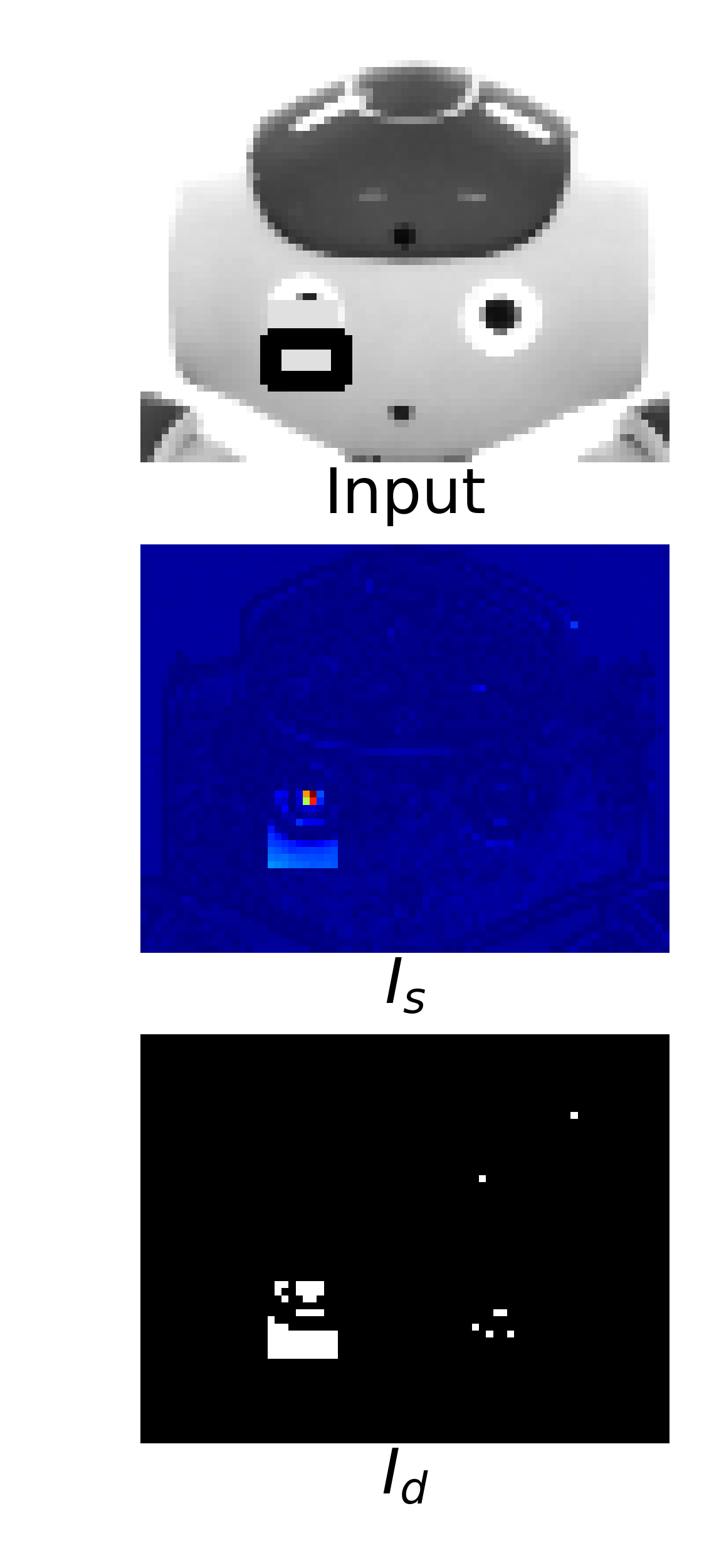}
		} 
		\subfigure[pink post-it]{			
			\includegraphics[width=0.15\textwidth]{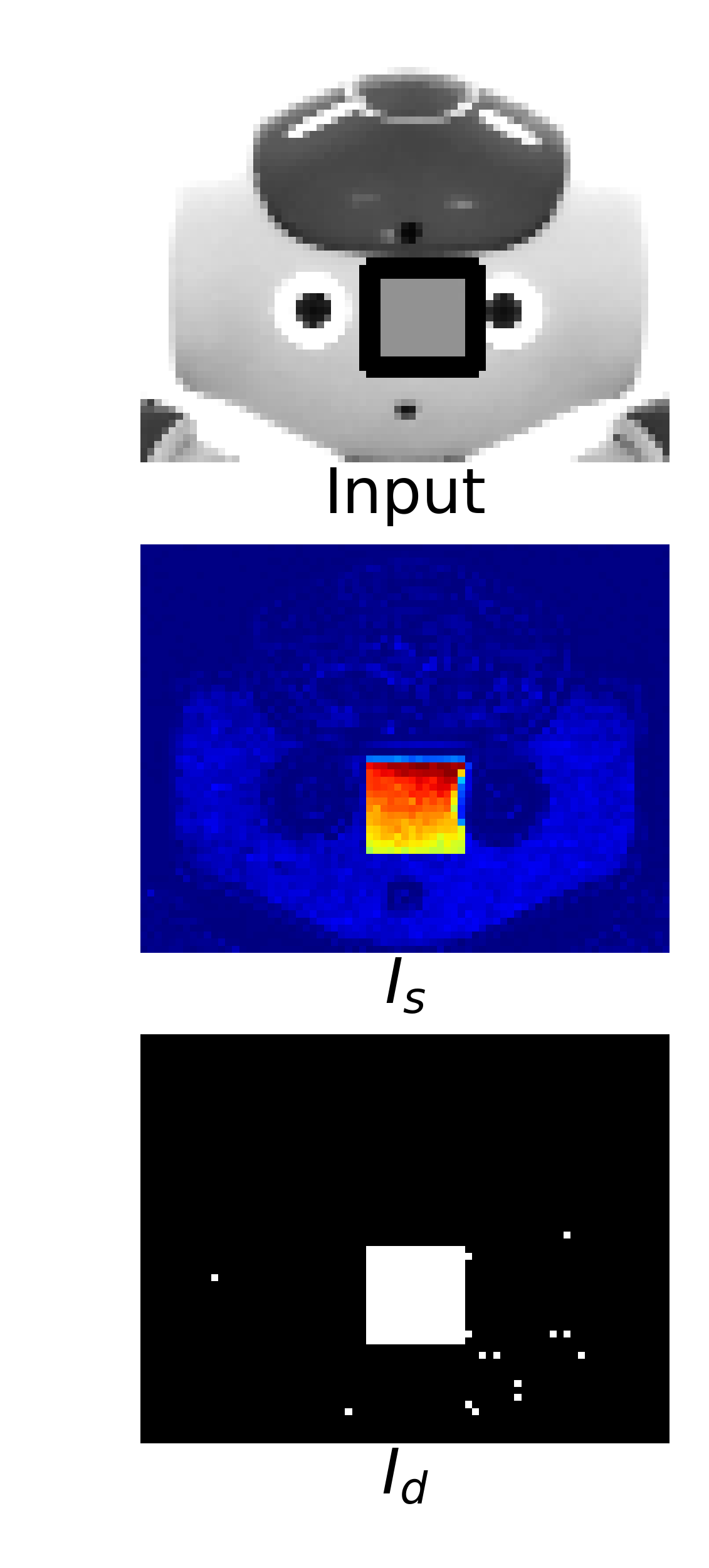}
		} 		
		\subfigure[bar mark]{
			\includegraphics[width=0.15\textwidth]{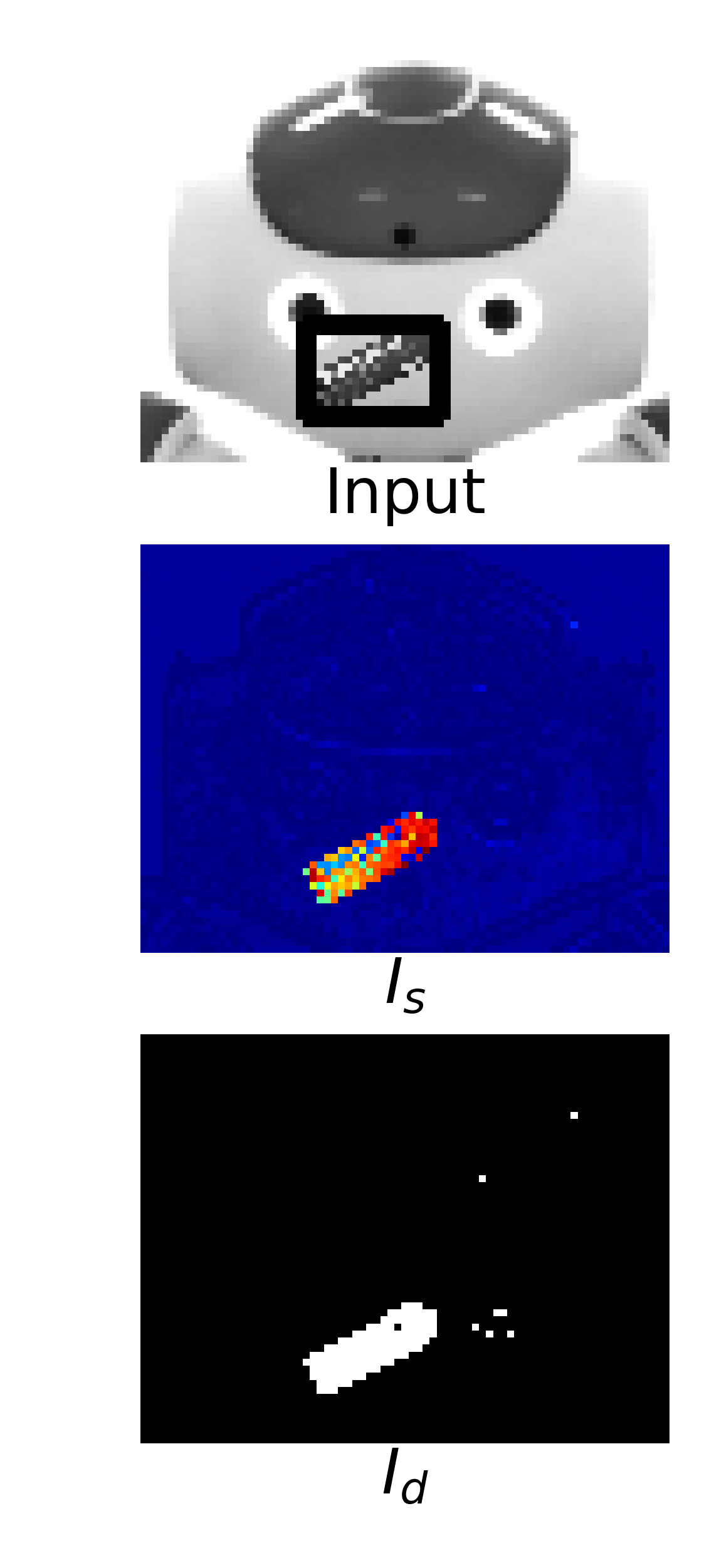}
		} 
		\subfigure[horizontal]{
		\includegraphics[width=0.15\textwidth]{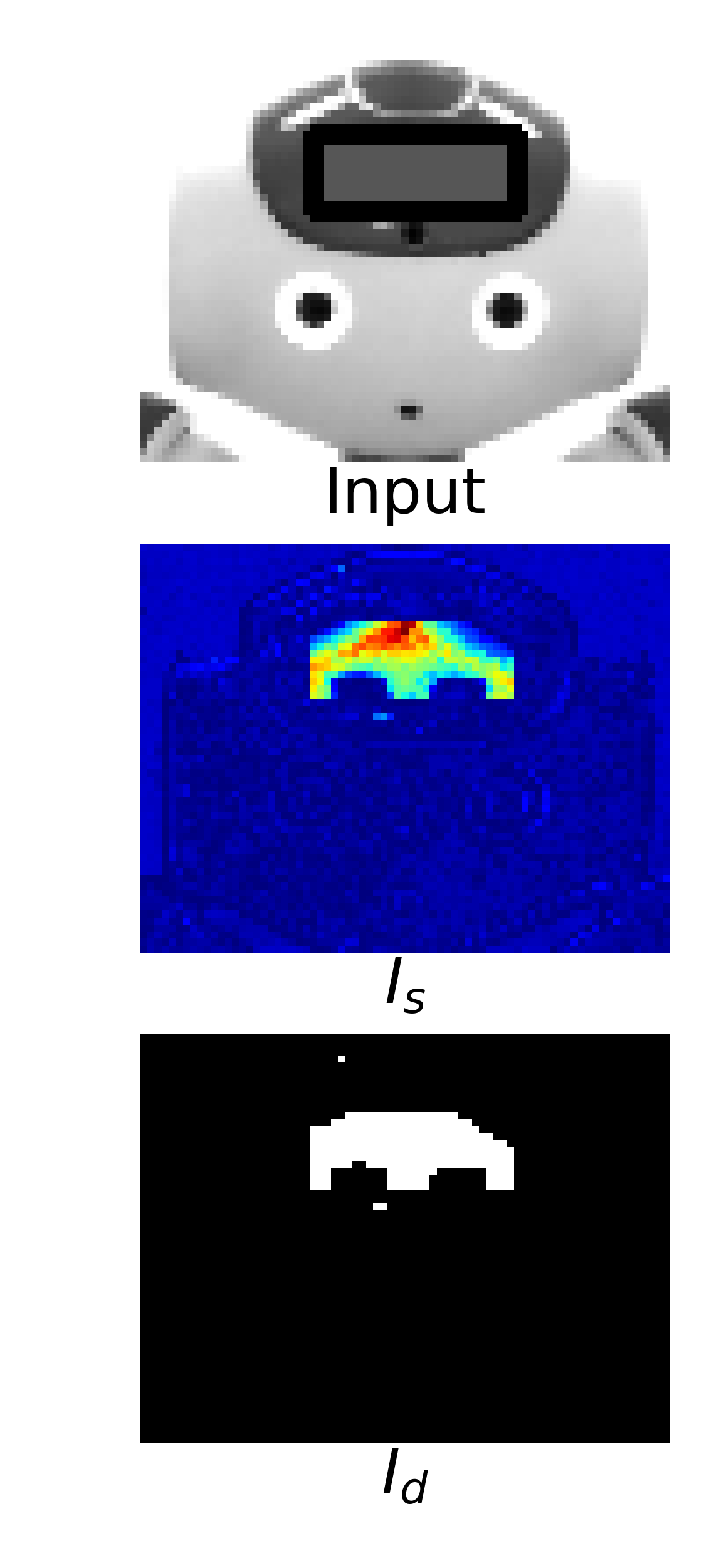}
	} 
		\caption{Saliency and mark detection examples. The input image is the automatically cropped face; $I_s$ is the computed saliency; $I_d$ is the binarized saliency, i.e., mark detection output.}
		\label{saliency}
	\end{figure*}

	Figure \ref{saliency} shows five examples of the saliency $I_s$ and the mark detection $I_d$ computation using different mark colors and shapes. Although the highest saliency region corresponded to the mark, we can observe other parts of the face that have relevance, such as the face boundaries. Eyes had a high variance in the prediction error, affecting the mark detection.
		
	\subsection{Online mirror test on real robots}
	We finally tested our architecture on the real Nao robots. On the ``standard'', red, robot (Fig. \ref{settingup} A-C), we first trained the face representation network with 800 real specular images. Second, we trained the visuo-proprioceptive mapping network (Section \ref{sec:reaching_for_mark}) by generating a database of visual mark locations and corresponding reaching joint states by manually moving the robot arm to reach close to the mark. To this end, we selected 5 joints\footnote{Elbow pitch and wrist rotation were fixed.} (\textit{Head yaw}, \textit{Head pitch}, \textit{Left shoulder pitch}, \textit{Left shoulder roll} and \textit{Left elbow roll}) as the output of the ANN and we manually moved the robot arm and head to reach the mark in the specular image. The reaching space of the robot is constrained to the face in the training phase, thus, placing the mark outside the face could result in interpolated or undesired reaching behaviors. ADAM optimizer was used for the training; the learning rate was set to 0.01 and the total training iterations were $10^4$.
	
\begin{figure*}[hbtp!]
		\centering
		\includegraphics[width=.9\textwidth]{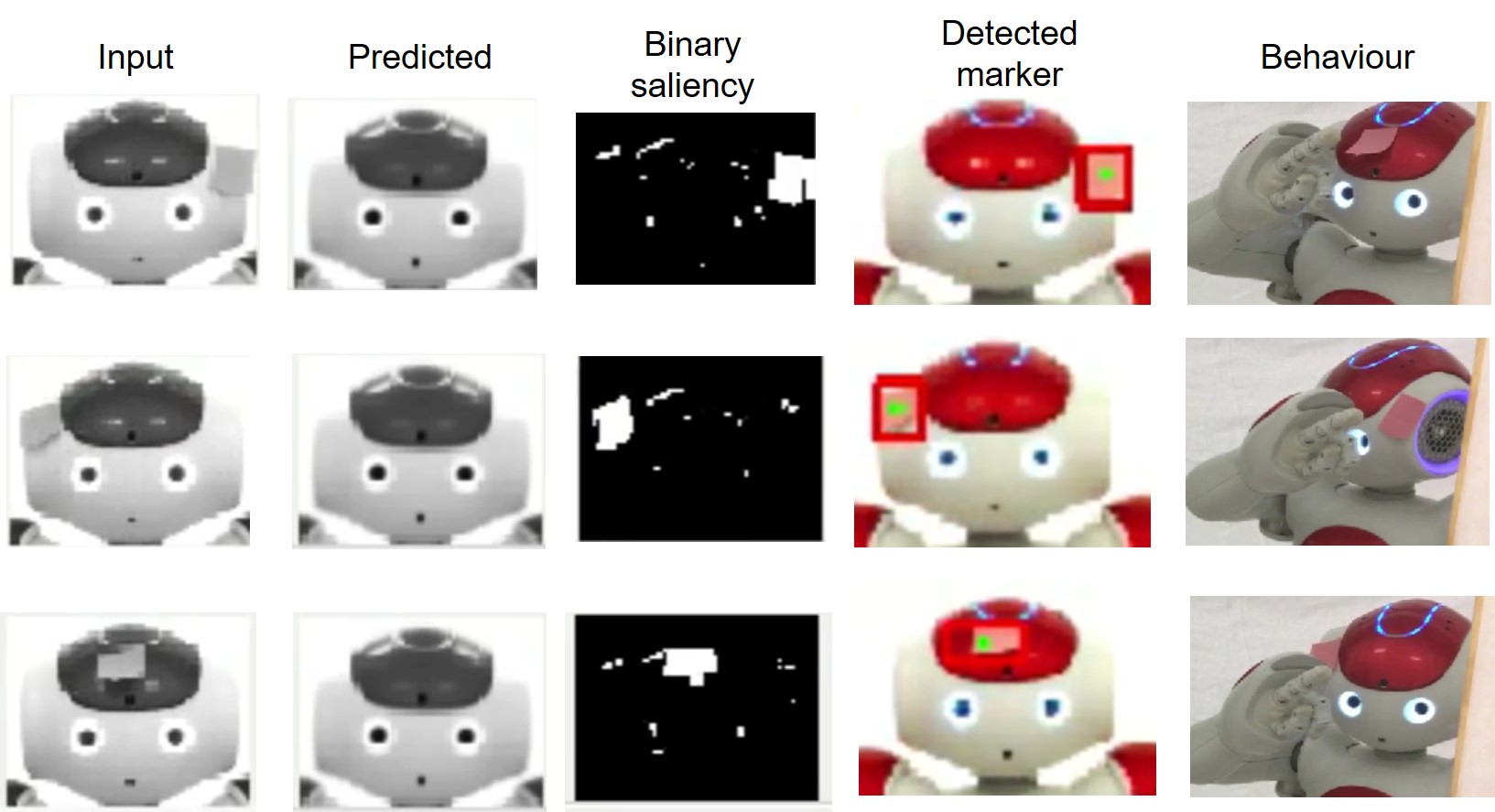}
		\caption{Examples of passing the mirror test with the Nao robot. The input corresponds to the input image to the ANN and the output is defined by the reaching behavior.}  \label{real-time}
	\end{figure*}

After training, post-it notes of different colors and sizes were used as marks and their position was changed after every reaching response. Figure \ref{real-time} shows examples from three trials. For these tests, we only used the decoder-like model as the generative process to predict the visual input.
	
\begin{figure*}[hbtp!]
		\centering
		\includegraphics[width=.8\textwidth]{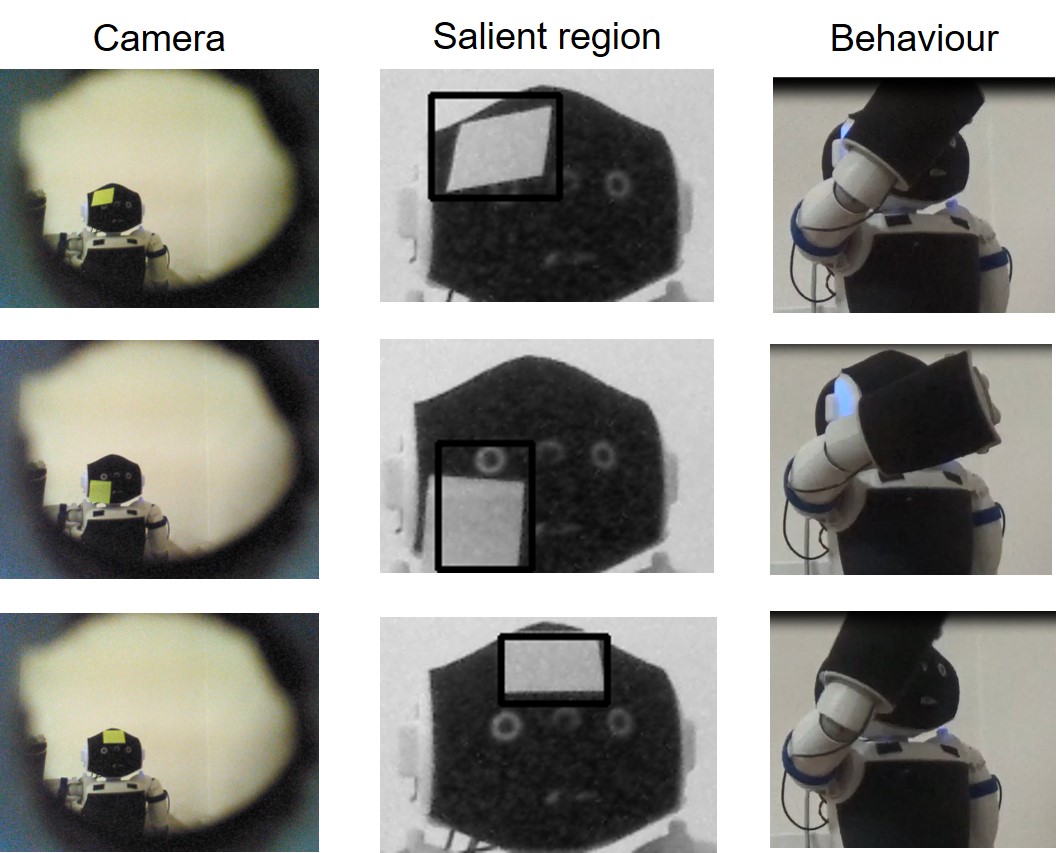}
		\caption{Examples of passing the mirror test with a second Nao robot with custom covers and pressure-sensitive skin covering the face.}  \label{real-time-touch}
	\end{figure*}

	 To verify the robustness of our approach, we implemented the same architecture in a second robot---with a different setup and mirror and using a robot with very different visual appearance. Training had to be repeated. Figure \ref{real-time-touch} shows three executions with the other Nao robot. The columns show the camera registered image, the detected mark, and the corresponding reaching behavior respectively.
	
	\section{DISCUSSION}
	\label{sec:discussion}
	    	
	The nature of our model is quite different than that of Mitchell \cite{mitchell1993mental}. The theories that Mitchell puts forth correlate success in MSR with other capacities: visual-kinesthetic matching, understanding mirror correspondence, object permanence, or objectifying body parts. However, testing each of these is, first, a challenge in itself. Second, such evidence still falls short of explaining the mechanisms of MSR. Instead, our modeling targets the process of undergoing MSR by developing an embodied computational model on a robot in front of a mirror. We do not treat MSR as a binary test---where one can succeed or fail and where success can be achieved (or engineered) in many ways---but as a process in which the behavioral manifestations of getting through the test can help us uncover the putative mechanisms. In this work, we model the mark detection on the face in some detail. Using novelty detection in the image of one's face as the behavioral trigger is an assumption we are making, inspired by the attention system studies in humans \cite{corbetta2002control}. The relation of our model to active inference \cite{friston2010free,apps2014free,lanillos2020robot} lies in its generative nature. The robot uses the internal representation to predict its appearance and the prediction error triggers the movement. However, our model does not yet account for generating the actions directly from prediction errors.
	
	%\added{Currently, we leave aside probabilistic, predictive coding or active inference accounts that constitute alternatives for the modeling endeavor \cite{apps2014free,rao1999predictive,lanillos2020robot}.}

	Reaching for the mark is currently engineered. However, we think that this very process of reaching is key to understanding MSR as it is the clearest response and can be analyzed quantitatively. We propose different ways of how the subjects may reach for the mark (Section~\ref{sec:mechanisms}, Fig.~\ref{fig:msr_mechanism_schematics}) and it is our plan to model these in the future. However, more information about how infants and other animals reach for the targets is needed.
			
    Variables that may be instrumental in understanding the mechanisms generating the response are:   
	\begin{itemize}
	    \item familiarization phase: do the subjects exploit temporal contingency between their movements and those of their reflection in the mirror to test that it is them in the mirror?
	    \item gaze / eye tracking: where do the subjects look: (i) target (mark on the face) in the mirror, (ii) alternate between target and their hand in the mirror, (iii) look at their hand directly
	    \item elements of tactile localization: after an initial inaccurate reaching for the mark, is touch used to bring the hand closer to the target?
	    \item movement duration: from mark placement to touching the mark
	    \item do subjects reach for the correct location but on the other side of their face?
	    \item reaching accuracy
	    \item repeated touches: do the subjects touch the mark repeatedly? Is there any exploration? 
	    \item are neck joints / head movements involved? do they assist mark localization or retrieval?
	    \item arm movement kinematics
	\end{itemize}
	Additional control experiments could involve: (1) reaching for other targets visible in the mirror or use of distorting mirrors to isolate whether the movements are visually guided on the mirror reflection; (2) reaching to the mark visible in the mirror could be contrasted with reaching for targets that can be perceived by other modalities---in isolation or together with visual perception through the mirror. Chang et al.~\cite{chang2015mirror} employed visual-somatosensory stimuli (high-power laser) in macaques; Chinn~\cite{chinn2019development} compared infants' reaching for vibrotactile target on the face away and facing the mirror with the rouge localization task; (3) with infants, one may instruct them also verbally to touch their nose and compare the reaching movements. Finally, one should keep in mind that the mismatch between how one's face normally looks like and the mirror reflection with the mark may not be what is tested. The mark is typically highly salient; additionally, the subject may also interpret the reflection as a conspecific with a mark on the face, which triggers a reaching response to check whether the mark may be also on her own face. One may thus be testing simply reaching for a target on the face with the help of a mirror. Learning the face representation and novelty detection needs also further investigation. While our own face reflection strongly produces attentional capture, the mark is even more salient. Disambiguation of self-face saliency and pure novelty can be investigated in a MSR setting looking at a non-self face in the mirror and under the face-illusion~\cite{ma2017creating}.

    \section{CONCLUSION AND FUTURE WORK}
    \label{sec:conclusion}
    In this work, we provided a mechanistic decomposition, or process model, of what components are required to pass the mirror recognition test. Second, we developed a model on a humanoid robot Nao that passes the test, albeit side-stepping some of the components needed by engineering them. The core of our technical contribution is learning the appearance representation and visual novelty detection by means of learning the generative model of the face with deep auto-encoders and exploiting the prediction error. 
    
	The proposed architecture uses a deep neural network to learn the face representation and subsequently deploys this as a novelty detector by exploiting the prediction error. The novelty detection network (autoencoder) is currently based on state of the art in machine learning and computer vision. To what extent this is compatible with the computation in the brain is debatable. Using more biologically realistic neural networks and learning algorithms is a direction of future research. Variance weighted prediction error was relevant to properly detect the mark in the face. Behavioral response---reaching for the mark---was achieved by learning a mapping from the salient region on the face to robot joint configuration required for the reach. The framework was quantitatively tested on synthetic data and in real experiments with different colored marks, sizes and shapes. Furthermore, two robots with completely different visual appearance were used for real-world testing.
	
	Although we were currently investigating how mainly one sensor modality (visual appearance data) produces the behavior, it is important to highlight that self-recognition is a multimodal process \cite{chang2015mirror,deWaal2019fish}. 
	In the future, we hope to acquire additional multimodal data (including movement kinematics and touch) about the details how infants or animals succeed in the mirror mark test and use these as constraints for our computational architecture---adding proprioception and touch. In particular, more information is needed to inform the process of reaching for the mark. One promising computational approach, which directly connects the error prediction scheme of the novelty detection with the action, would be active inference goal driven control~ \cite{oliver2019active,sancaktar2020end,rood2020deep}, where the visual error would produce a reactive reaching behavior toward the mark.

	%                                   % the textheight of the last page by a suitable amount.
	%                                   % This command does not take effect until the next page
	%                                   % so it should come on the page before the last. Make
	%                                   % sure that you do not shorten the textheight too much.
	
	% %%%%%%%%%%%%%%%%%%%%%%%%%%%%%%%%%%%%%%%%%%%%%%%%%%%%%%%%%%%%%%%%%%%%%%%%%%%%%%%%

	% %%%%%%%%%%%%%%%%%%%%%%%%%%%%%%%%%%%%%%%%%%%%%%%%%%%%%%%%%%%%%%%%%%%%%%%%%%%%%%%%


	
	% % \section*{ACKNOWLEDGMENT}
	
	% % The preferred spelling of the word “acknowledgment” in America is without an “e” after the “g”. Avoid the stilted expression, “One of us (R. B. G.) thanks . . .”  Instead, try “R. B. G. thanks”. Put sponsor acknowledgments in the unnumbered footnote on the first page.

	% %%%%%%%%%%%%%%%%%%%%%%%%%%%%%%%%%%%%%%%%%%%%%%%%%%%%%%%%%%%%%%%%%%%%%%%%%%%%%%%%
	
	% References are important to the reader; therefore, each citation must be complete and correct. If at all possible, references should be commonly available publications.

	\bibliographystyle{plain}
	\bibliography{pl,references}

\end{document}